# Application of Deep Learning in Neuroradiology: Automated Detection of Basal Ganglia Hemorrhage using 2D-Convolutional Neural Networks.

Vishal Desai,[1] Adam E. Flanders,[1] Paras Lakhani[1*]

1. Department of Radiology, Thomas Jefferson University Hospital, Sidney Kimmel Jefferson Medical College, Philadelphia, PA, U.S.A.
   *Corresponding Author, email: Paras.lakhani@jefferson.edu

*Abstract— Purpose:* **Deep learning techniques have achieved high accuracy in image classification tasks, and there is interest in applicability to neuroimaging critical findings. This study evaluates the efficacy of 2D deep convolutional neural networks (DCNNs) for detecting basal ganglia (BG) hemorrhage on noncontrast head CT.** *Methods:* **170 unique de-identified HIPAA-compliant noncontrast head CTs were obtained, those with and without BG hemorrhage. 110 cases were held-out for test, and 60 were split into training (45) and validation (15), consisting of 20 right, 20 left, and 20 no BG hemorrhage. Data augmentation was performed to increase size and variation of the training dataset by 48-fold. Two DCNNs were used to classify the images—AlexNet and GoogLeNet—using untrained networks and those pre-trained on ImageNet. Area under the curves (AUCs) for the receiver-operator characteristic (ROC) curves were calculated, using the DeLong method for statistical comparison of ROCs.** *Results:* **The best performing model was the pre-trained augmented GoogLeNet, which had an AUC of 1.00 in classification of hemorrhage. Preprocessing augmentation increased accuracy for all networks (p<0.001), and pretrained networks outperformed untrained ones (p<0.001) for the unaugmented models. The best performing GoogLeNet model (AUC 1.00) outperformed the best performing AlexNet model (AUC 0.95)(p=0.01).** *Conclusion:* **For this dataset, the best performing DCNN identified BG hemorrhage on noncontrast head CT with an AUC of 1.00. Pretrained networks and data augmentation increased classifier accuracy. Future prospective research would be important to determine if the accuracy can be maintained on a larger cohort of patients and for very small hemorrhages.**

*Index Terms—*deep learning, intracranial hemorrhage, basal ganglia, convolutional neural networks

## I. INTRODUCTION

Basal ganglia hemorrhage is a type of intracerebral hemorrhage (ICH), and is associated with long-standing hypertension.[1–3] Along with other forms of ICH, it is considered a neurologic emergency. The imaging modality of choice to detect such hemorrhages is non-contrast head CT given its wide availability, speed at which it can be performed, and the very high sensitivity and specificity for detecting acute hemorrhage.[4]

The American Heart Association and American Stroke Association note that timely diagnosis and aggressive early management is very important in ICH, as affected patients commonly deteriorate within the first few hours after onset.[5]

As such, an automated solution to identify such hemorrhages may be helpful to decrease time to diagnosis, and more readily triage appropriate care. Prior work with computer aided detection (CAD) has shown success in automated segmentation of ICH.[6] In addition, one study using CAD achieved 95% sensitivity and 89% specificity for detection of ICH using pre-processing techniques and a knowledge-based classification system.[7] Another study using machine learning techniques was able to achieve 96% accuracy for detecting ICH using pre-processing techniques, feature selection, and a fuzzy classifier.[8]

Based on the recent success in the ImageNet Large Scale Visual Recognition Competition, deep convolutional neural networks (DCNNs) are considered state-of-the-art for image classification.[9] While convolutional neural networks have been around for many years, only recently have we seen their increase used due to availability of high-performing graphics processing units (GPU) and availability of many open-source frameworks. Since 2012, deep convolutional neural networks (DCNN) have been utilized by all winning entries in ILSVRC, resulting in a drop in top-5 classification error rate from 25% in 2011 to 3% more recently with some of the latest architectures.[10]

Application of deep learning in radiology has found promising results in multiple modalities. Some recent examples include brain segmentation on MRI,[11] pancreatic segmentation,[12] knee cartilage evaluation,[13] detection of pleural effusion and cardiomegaly on chest radiographs,[14] detection of tuberculosis on chest radiographs,[15] and intracranial critical findings.[16]

In this study, we focus on identifying a common site of hemorrhagic stroke, the basal ganglia. We evaluate the efficacy of automated detection of basal ganglia hemorrhage on noncontrast head CTs using two DCNNs – AlexNet[17] and GooLeNet,[18] the winners of the 2012 and 2014 ImageNet

Large Scale Visual Recognition Competition respectively.

II. METHODS

*Datasets:*

This was a retrospective study using de-identified and Health Insurance Portability and Accountability Act (HIPAA) compliant datasets from Thomas Jefferson University Hospital, PA, U.S.A. To create the datasets, the institutional Radiology Information System (RIS) database was searched over a 12-month period between May 2016 - May 2017 for patients with basal ganglia hemorrhage and for patients without any acute intracranial pathology on non-contrast head CT. All CT studies were verified by two board-certified radiologists, the author of the final report, and by an independent board-certified radiologist (P.L.). The terms "basal ganglia, hemorrhage, hematoma, and bleed" were entered into the database using multiple permutations. Patients with brain tumors and prior cranial surgery were excluded.

170 unique patients were identified. 110 of these patients were held out and placed in the test dataset, consisting of 55 with BG hemorrhage, and 55 without hemorrhage. The remaining 60 patients were used for training and validation, divided in a 75/25 ratio, with 45 cases for training and 15 cases for validation. Of these 60 training/validation cases, 20 consisted of right basal ganglia hemorrhage (15 training + 5 validation), 20 left basal ganglia hemorrhage (15 training + 5 validation), and 20 no intracranial hemorrhage (15 training + 5 validation). Seven cases with basal ganglia calcifications (mimic of BG hemorrhage) were included in the datasets, four for training and 3 in the test-dataset. The training set was used to train the algorithm, validation set for model selection, and test set for assessment of the final chosen model. 110 test cases were chose to provide a 95% confidence interval of ±7.5% based on estimated accuracy of the best-performing models on the accuracy on the validation datasets.[19]

*Study Design:*

For each head CT in the training dataset, several axial key images were obtained near or at the level of the basal ganglia, which were determined by two radiologists (V.D., P.L.). This sometimes included images one slice above and/or below the basal ganglia, and approximately 3-5 images at the level of the basal ganglia. Approximately 4-6 images were saved per study depending on the scan angle, slice thickness (range 2.5-5 mm), and, if present, size of the hemorrhage. A total of 308 unique DICOM (Digital Imaging and Communications in Medicine) images at the levels of interest were obtained from 60 non-contrast head CTs (20 with right basal ganglia hemorrhage, 20 with left basal ganglia hemorrhage, and 20 without hemorrhage). The images were resized to 256 x 256 pixels (from 512 x 512 originally) and converted into Portable Network Graphics (PNG) format. The images were loaded onto a workstation running the DIGITS deep learning GPU training system (DIGITS 4.0, Nvidia Corporation, Santa Clara, CA) running Ubuntu 14.04, Caffe deep learning framework (Nvidia fork), CUDA 8.0, and cuDNN dependencies (Nvidia Corporation, Santa Clara, CA) for graphics processing unit (GPU) acceleration. The computer contained an Intel i5 3570k 3.4gHz processor, 4TB hard disk space, 32gb RAM, and a CUDA-enabled NVIDIA Maxwell Titan X 12Gb GPU (Nvidia Corporation, Santa Clara, CA).

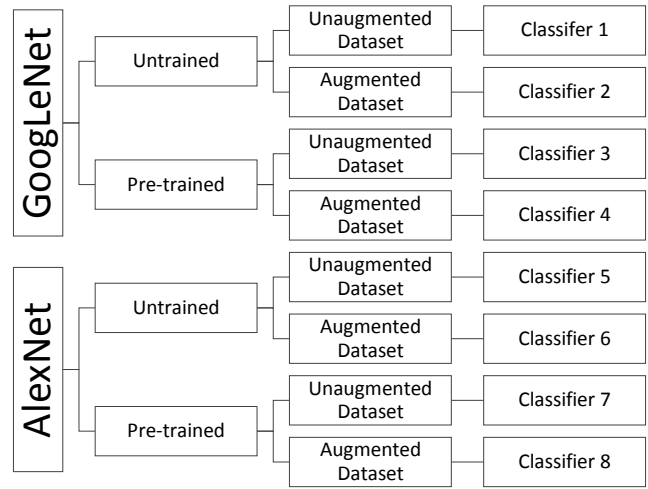

**Figure 1:** Eight different models were evaluated, using two different architectures (AlexNet and GoogLeNet), including pretrained and untrained networks, and with and without extra augmentation.

Multiple sequential pre-processing augmentation techniques were performed, including three different CT window-width (WW) and window-levels (WL) ( "brain" 80/35 WW/WL, "acute blood" 220/95 WW/WL, and "stroke" 27/33 WW/WL), 10% magnification, rotation of 15 & 30 degrees, blur and edge enhancement. This pre-processing was performed using ImageJ v1.50i (National Institutes of Health, U.S.A.) and XnConvert v1.73 (XnSoft, France). Images were also augmented in real-time during the training process using prebuilt options within the Caffe framework,[16] including random cropping of 227 x 227 pixels for AlexNet and 224 x 224 pixels for GoogLeNet, and mean whole-image subtraction. After image augmentation using pre-processing detailed above, the new training dataset totaled 11,088 images, which was a 48-fold increase in number of images from the originals.

Two deep convolutional neural networks architectures were used—AlexNet and GoogLeNet, using unaugmented and augmented datasets, and those untrained and pre-trained on ImageNet. Taking the permutation of the two datasets (unaugmented and augmented datasets), two neural networks (AlexNet and GoogLeNet), and two neural network types (pre-trained and untrained), a total of 8 trained models were created (**Figure 1**). Pre-trained networks were obtained from the Caffe Model Zoo (http://caffe.berkeleyvision.org), Berkeley, CA), an open-access repository of pre-trained models for use with Caffe,[20] and were previously trained on over 1 million every-day color images from ImageNet.[9]

The following solver parameters were used for training AlexNet: 90 epochs, base learning rate of 0.01 for untrained models and 0.001 for pre-trained models, stochastic gradient descent, step-down 33%, and gamma of 0.5. For GoogLeNet, the parameters were: 40 epochs, base learning rate of 0.01 for untrained models and 0.001 for pre-trained models, stochastic gradient descent, step-down 33%, and gamma of 0.5. Categorical cross-entropy was used for the loss function. The solver parameters including number of epochs were determined after reviewing the training and validation loss curves after multiple runs of the data, with the goal of achieving the lowest validation loss, and stopping training after plateau in the loss. For pretrained networks, we randomized the weights of the final fully-connected layer of the networks to learn from the CT images, and employed a fine-tuning strategy where all layers were open to learn at a reduced base learning rate.

*Statistical Analysis:*

Receiver operating characteristic (ROC) curves and area under the curves (AUC) were determined using the pROC package (ver. 1.7.3) for R (ver. 3.3.1), utilizing the DeLong method for statistical comparison of ROCs.[21,22] P-values < 0.05 were considered statistically significant.

*IRB approval:*

This study was approved by the Institutional Review Board at Thomas Jefferson University Hospital, Philadelphia, PA, U.S.A.

## III. RESULTS

The average size of the hemorrhages for the test dataset was 2.8±1.9cm in the transverse dimension, and 3.6±2.7 cm in the anteroposterior dimension. The smallest two hemorrhages were 0.5 x 1.1cm and 0.7 x 0.7cm, and the largest was 7.1 x 11.2cm.

**Table 1** contains area under the curve (AUC) calculations for all classifiers. The classifiers trained with the augmented dataset had significantly better AUC than those trained with the smaller, unaugmented dataset in all scenarios–pre-trained GoogLeNet and AlexNet and untrained GoogLeNet and AlexNet (Figure 2 & 3, Table 2). The pre-trained classifier was significantly better than the untrained classifier for GoogLeNet when using the unaugmented dataset (Table 3). There was no significant difference in AUC between the untrained and pretrained GoogLeNet and AlexNet classifiers when using the augmented dataset (Table 4).

The best performing classifiers were the augmented untrained and pre-trained GoogLeNet DCNN, which had an AUC of 0.99 and 1.0 respectively in identifying BG hemorrhage (Table 1). The augmented GoogLeNet classifiers correctly identified 55/55 cases of BG hemorrhage in the test dataset, including the laterality of the bleed (sensitivity 100%). The pre-trained classifier correctly labeled 55/55 cases without hemorrhage (specificity 100%) while the untrained classifier had one false positive in a case with asymmetric basal ganglia calcifications (specificity 96.4%).

The augmented untrained AlexNet classifier retained a high AUC (0.96), with sensitivity for hemorrhage at 100% but with a lower specificity at 80%, particularly mislabeling basal ganglia calcifications as hemorrhages (Figure 4). Overall, the augmented pre-trained GoogLeNet classifier had greater accuracy compared to the AlexNet classifiers (Table 5, Figure 3). The best performing GoogLeNet model (AUC 1.00) outperformed the best performing AlexNet model (AUC 0.95)(p=0.01).

**Table 1: AUC for All Classifiers**

| Classifier | Area Under the Curve |
|---|---|
| Unaugmented AlexNet-Untrained | 0.57 (0.46-0.68) |
| Unaugmented AlexNet-Pretrained | 0.81 (0.74-0.89) |
| Unaugmented GoogLeNet-Untrained | 0.60 (0.49-0.70) |
| Unaugmented GoogLeNet-Pretrained | 0.89 (0.83-0.95) |
| Augmented AlexNet-Untrained | 0.95 (0.92-0.99) |
| Augmented AlexNet-Pretrained | 0.92 (0.87-0.97) |
| Augmented GoogLeNet-Untrained | 0.99 (0.96-1.00) |
| **Augmented GoogLeNet-Pretrained** | **1.00 (1.00-1.00)** |

Parentheses reflects 95% confidence interval

## IV. DISCUSSION:

Recent advances in technology have enabled the use of machine learning, specifically deep learning, to be applied to radiologic images. Machine learning allows computers to analyze data and perform tasks without being explicitly programmed to do so.[23] Deep artificial neural networks, on the other hand, is a relatively newer branch of machine learning, which consists of multiple hidden layers, and excels with high dimension datasets such as images. For this study, we used supervised (pre-labeled) images to train our basal ganglia hemorrhage classifiers. DCNNs consist of a varying number of interconnected layers, and each layer contains numerous independent "nodes" or "neurons" which analyze a certain feature. Since all layers are interconnected, this creates a network where each data point is constantly referenced against all other layers to provide the best possible prediction.

**Table 2: AUC Comparison of Classifiers with Unaugmented and Augmented Datasets**

| Classifier | Unaugmented AUC | Augmented AUC | p-value |
|---|---|---|---|
| AlexNet-Untrained | 0.57 | 0.95 | p<0.001 |
| AlexNet-Pretrained | 0.82 | 0.92 | p=0.001 |
| GoogLeNet-Untrained | 0.60 | 0.99 | p<0.001 |
| **GoogLeNet-Pretrained** | 0.89 | **1.00** | p<0.001 |

Deep neural networks have been described as "black boxes," where the reasoning behind a network's prediction may not be readily apparent.[24] For medical imaging, ensuring the network has been properly trained and performs accurately in unknown cases is crucial. The black box scenario is compounded by the fact that DCNNs are extremely large and complex.

**Table 3: AUC Comparison of Untrained and Pre-trained Classifiers with Unaugmented Dataset**

| Classifier | Untrained AUC | Pre-trained AUC | p-value |
|---|---|---|---|
| AlexNet, Unaugmented | 0.57 | 0.81 | p<0.001 |
| GoogLeNet, Unaugmented | 0.60 | 0.89 | p<0.001 |

However, several strategies do exist to get insight into what is contributing to the network's prediction.[24-26] One technique is to obscure the area of interest and then assess how this affects the network's prediction.[25] For instance, obscuring a basal ganglia hematoma by overlaying a gray box should result in a prediction of "no basal ganglia hemorrhage" (if the network has learned what was intended) (Figure 5). Rather than manually processing images, we added laterality to our classifier, which is a simple method to verify appropriate training. So, if the classifier is erroneously labeling right basal ganglia hemorrhages as left, we can assume that the network is inappropriately trained.

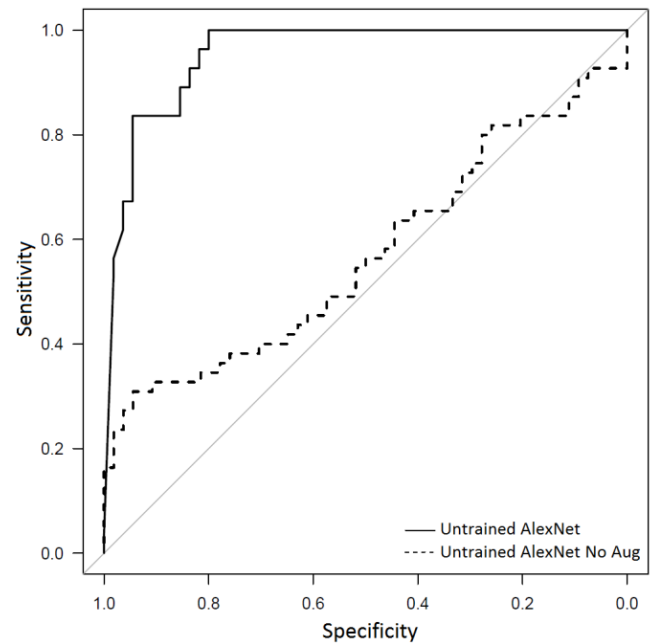

**Figure 2.** Receiver operating characteristic (ROC) curve comparing the Untrained AlexNet with and without extra augmentation, with an AUC of 0.95 and 0.57 respectively, p<0.0001.

Activations within a layer can also offer confidence that the network is being activated by the area of interest.[26] Figure 6 demonstrates a strong activation in 2nd convolutional layer of the pretrained GoogLeNet classifier followed by a correct prediction of left basal ganglia hemorrhage with 100% confidence.[26]

One helpful step in building an accurate image classifier is augmentation.[27] The results of our study demonstrate the statistically significant difference in AUCs among the unaugmented and augmented datasets (Table 2). It has been demonstrated that the more cases and variations supplied to the neural network during training, the better the performance and generalization of the DCNN.[27]

For the basal ganglia classifier, we ensured variety in the initial, unaugmented dataset by including cases with basal ganglia calcifications, different sized hemorrhages, multi-compartmental hemorrhage, and hydrocephalus. Then, with image augmentation we take that variety of cases and expose it to the neural network in different ways by introducing varying degrees of blur, edge enhancement, rotation, zoom, windowing and other image processing techniques. This simulates variations in the CT scan that might be encountered on a day-to-day basis, improving generalization of our classifier without needing to acquire extra cases. A degree of manual input is certainly required in the case layers for AlexNet) and employs an "inception" module with a "bottleneck" design.[18,28] The inception module uses smaller 1x1 convolutions to reduce the number of variables that flow into the layer, before larger convolutions, which improves computational efficiency and may improve predictions.[28]

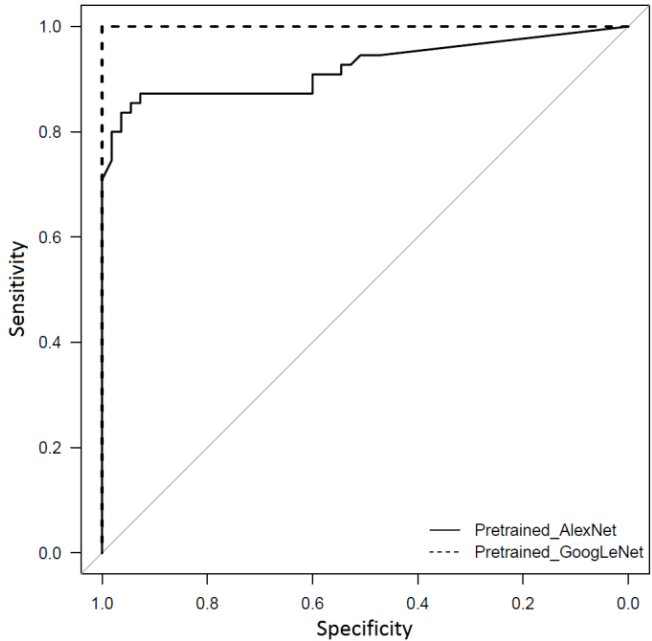

**Figure 3:** ROC curves of the pretrained AlexNet and GoogLeNet models with extra augmentation. The best-performing pretrained GoogLeNet model (AUC 1.00) performed better than the corresponding AlexNet model (AUC 0.92), p=0.006.

**Table 5: AUC Comparison of GoogLeNet and AlexNet with Augmented Dataset**

| Classifier | AlexNet | GoogLeNet | p-value |
|---|---|---|---|
| Untrained, Augmented AUC | 0.95 | 0.99 | p=0.12 |
| **Pretrained, Augmented AUC** | 0.92 | **1.00** | p=0.006 |

We also found that with small (unaugmented) datasets, the networks pre-trained with everyday images on ImageNet were superior to untrained networks. However, with larger (augmented) datasets, there was no significant difference between pre-trained and untrained networks (Table 4). Since all images, including medical images, share basic low-level features such as edges, lines and blobs, a network pre-trained on non-medical images can help prime the initial layers of the network through transfer learning.[29] Then, the interconnected layers could re-learn from the CT images provided. This appeared less significant with large datasets, perhaps because there was enough input data to robustly train the initial layers without pretraining.

selection process. For instance, we chose to specifically include cases with basal ganglia calcifications to decrease our false positive rate. The augmented GoogLeNet classifiers successfully labeled basal ganglia calcifications as "no hemorrhage" while AlexNet had difficulty with dense asymmetric calcifications, possibly related to the extra depth of the classifier.

**Table 4: AUC Comparison of Untrained and Pre-trained Classifiers with Augmented Dataset**

| Classifier | Untrained AUC | Pre-trained AUC | p-value |
|---|---|---|---|
| AlexNet, Augmented | 0.95 | 0.92 | p=0.27 |
| **GoogLeNet, Augmented** | 0.99 | **1.00** | p=0.22 |

GoogLeNet was superior to AlexNet using the augmented dataset (Table 5), although only statistically significant for the pre-trained models. This is likely due to the fact that GoogLeNet is a much deeper network (22 layers versus 8

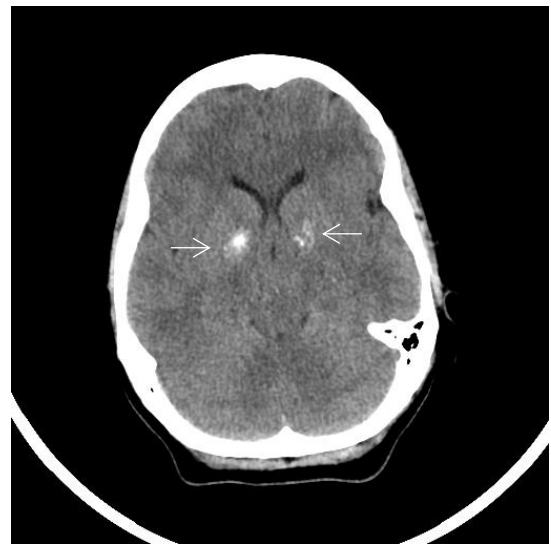

**Figure 4:** Axial noncontrast CT through the basal ganglia region (ww/wl: 80/35) demonstrating bilateral basal ganglia calcification (white arrows). This image was incorrectly predicted as hemorrhage by all classifiers except for the best-performing pretrained, augmented GoogLeNet model.

One of the common problems with deep learning is overfitting, or more simply, lack of generalizability.[30] In this scenario, the classifier maintains high prediction accuracy on the training dataset but does poorer with unknown cases, which can happen when the training dataset is small. The models employed several strategies to help overcome this, including increase in the size and variety of the dataset by pre-processing augmentation, and use of dropout—model regularization technique—which has been shown to help with overfitting.[27,30]

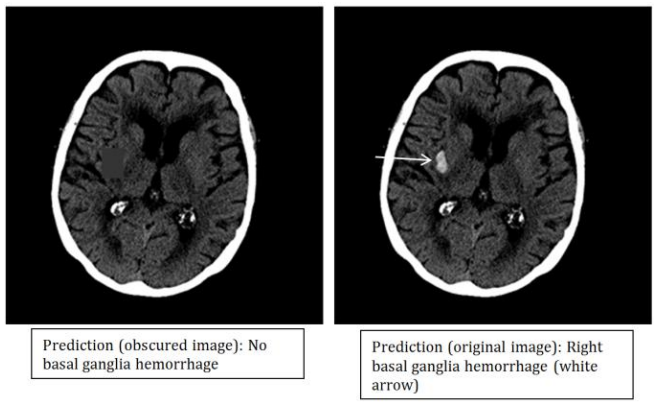

**Figure 5:** Axial non-contrast CT through the basal ganglia region (ww/wl: 220/95). The right image shows an acute hemorrhage in the right basal ganglia (white arrow), which is correctly predicted by the classifier (100% confidence). In the left image, the hemorrhage is obscured by a gray box, which changes the prediction to "no hemorrhage." This technique can help demonstrate that the model is assessing the appropriate region.

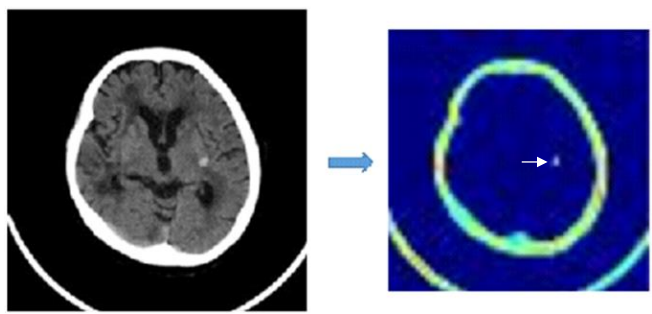

**Figure 6:** On the left is an axial CT image at the level of the basal ganglia (WW/WL: 80/35) containing a left-sided BG hemorrhage. On the right is an accompanying activation from the 2nd layer of the best-performing GoogLeNet model. The small orange-white focus (white arrow) represents an area activated by the network, and corresponds to the location of the hemorrhage.

Finally, we assess the trained models with test cases, previously unseen by the classifiers, and analyze the predictions (Table 1). This shows that the models can be generalizable to new unforeseen cases.

There are limitations to this study. While the image classifier we generated serves as a great proof-of-concept that deep learning can accurately detect basal ganglia hemorrhage, more research is needed to demonstrate its accuracy with larger cohorts. It is unknown how well the model would do in scenarios where there is significant artifact, prior cerebrovascular conditions such as brain tumors, or altered anatomy from prior surgery. In addition, it is unclear how sensitive the algorithm would be on more subtle subcentimeter hemorrhages. Most of the hemorrhages were greater than 1.5cm, although the classifier did detect some smaller hemorrhages measuring approximately 1cm (10 of the 55 positive test cases), including one that was 0.5 x 1.1cm and another that was 0.7 x .7cm. Also, this study was performed retrospectively and the performance of this classifier prospectively would need to be determined. Finally, this was a model that used 2D convolutions, and makes an assessment on a slice-by-slice basis. More research is needed to determine efficacy compared to 3D CNNs, which would have the benefit of analyzing the whole head but at greater computational costs and GPU memory needs. Before utilization in a clinical scenario, the classifier will need be trained to learn the normal anatomy of all parts of the cranium including the vertex and skull bases, rather than just the basal-ganglia region, so as not to flag normal structures as hemorrhage.

Despite the limitations, this study establishes a starting point for building an accurate CT-based image classifier for hemorrhage. The next steps will be to build a more-encompassing intracranial hemorrhage algorithm, and address the questions or limitations noted above.

Future work may also include a system where a deep learning algorithm automatically flags studies as positive or negative on a reading worklist, where one could then evaluate its effect on work prioritization and result turn-around-time.

## V. CONCLUSIONS:

For this dataset, the best performing DCNN identified BG hemorrhage on noncontrast head CT with an AUC of 1.00. Pretrained networks and data augmentation increased classifier accuracy. Future prospective research would be important to determine if high accuracy can be maintained on a larger cohort of patients and for very small hemorrhages.

## VI. REFERNCES: